%% file: main.tex
\pdfoutput=1
\documentclass[10pt,twocolumn,letterpaper]{article}

\usepackage[pagenumbers]{wacv}

\usepackage{graphicx}
\usepackage{amsmath}
\usepackage{amssymb}
\usepackage{booktabs}
\usepackage{placeins}
\input{preamble}

\usepackage[breaklinks,colorlinks]{hyperref}

\usepackage[capitalize]{cleveref}
\crefname{section}{Sec.}{Secs.}
\Crefname{section}{Section}{Sections}
\Crefname{table}{Table}{Tables}
\crefname{table}{Tab.}{Tabs.}

\title{AURASeg: Attention-Guided Upsampling with Residual-Assisted Boundary Refinement for Drivable-Area Segmentation}

\author{Narendhiran Vijayakumar}

\hypersetup{
  pdfauthor={Narendhiran Vijayakumar},
  pdftitle={AURASeg: Attention-Guided Upsampling with Residual-Assisted Boundary Refinement for Drivable-Area Segmentation},
  pdfsubject={arXiv preprint},
  pdfkeywords={drivable-area segmentation, boundary refinement, attention-guided decoding, embedded profiling}
}

\begin{document}
\maketitle

\input{sec/0_abstract}
\input{sec/1_intro}
\input{sec/2_formatting}
\input{sec/3_finalcopy}
\input{sec/4_results}
\input{sec/5_conclusion}

\FloatBarrier
{\small
  \bibliographystyle{ieee_fullname}
  \bibliography{main}
}

\end{document}

%% file: preamble.tex
\usepackage{microtype}
\usepackage{tabularx}
\usepackage{multirow}
\usepackage{pifont}

\newcommand{\cmark}{\ding{51}}

%% file: sec/0_abstract.tex
\begin{abstract}
Free-space segmentation is essential for autonomous
robots to identify drivable regions and navigate safely across
indoor, outdoor, and road-scene environments. However,
conventional encoder–decoder models often recover coarse
region masks while losing the fine spatial information needed
to localize drivable-area boundaries accurately. We propose Attention-Guided Upsampling with Residual-Assisted
Boundary Refinement (AURASeg), a segmentation framework designed to preserve region-level accuracy while improving boundary quality. Built on a ResNet-18 encoder,
AURASeg introduces an Attention Progressive Upsampling
Decoder (APUD) that progressively combines semantic context with high-resolution spatial detail, together with a Residual Boundary Refinement Module (RBRM) that explicitly
refines contour-sensitive features before final prediction. We
evaluate AURASeg across indoor simulation, ground-robot
imagery, and road-driving benchmarks. The results show
that our proposed model remains competitive with established segmentation models on region-level metrics while
providing particularly strong boundary localization, including in comparison with boundary-focused methods. Detailed
ablations further demonstrate the role of the proposed decoding and refinement modules.
\end{abstract}

%% file: sec/1_intro.tex
\section{Introduction}
\label{sec:intro}

\begin{figure}[!t]
  \centering
  \includegraphics[width=0.98\columnwidth]{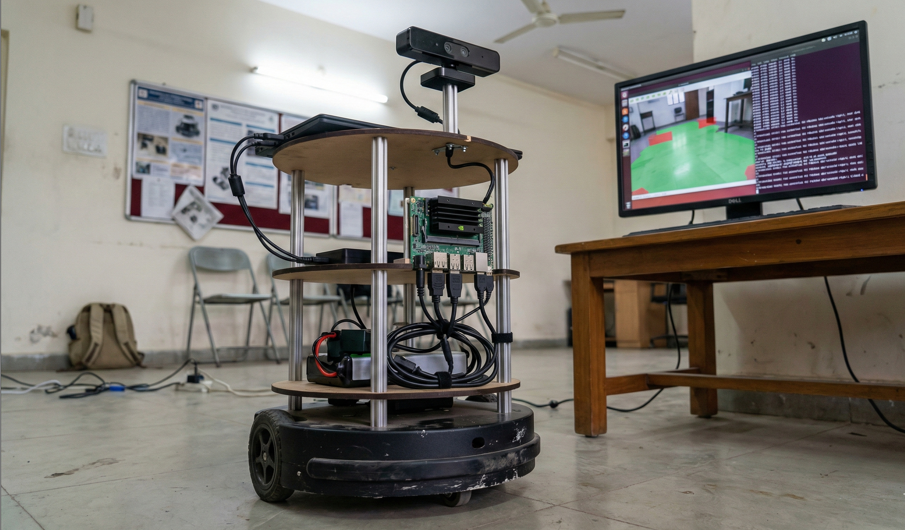}
  \caption{Onboard drivable-area segmentation on a Kobuki TurtleBot2 equipped with an NVIDIA Jetson Nano.}
  \label{fig:1}
\end{figure}

Autonomous robotic navigation relies on semantic segmentation to distinguish traversable ground from obstacles and support safe motion planning. Although modern segmentation networks provide strong region-level accuracy, feature downsampling and coarse decoder fusion can blur thin structures and floor--obstacle transitions. These errors are particularly important for free-space perception: a small contour displacement can remove navigable pixels or absorb obstacle pixels into the predicted drivable region. A useful drivable-area model must therefore preserve global region consistency while recovering precise local boundaries. The onboard robotic setting motivating this study is illustrated in \cref{fig:1}.

Multi-scale context aggregation can improve scene-level understanding, while efficient architectures such as BiSeNet~\cite{c4}, TwinLiteNet~\cite{c13}, and PIDNet~\cite{c28} retain higher-resolution information for dense prediction. Boundary-oriented methods further introduce refinement branches, auxiliary edge supervision, or explicit detail streams~\cite{c3,c15,c30}. These directions are complementary, but repeatedly mixing semantic, detail, and edge features throughout the network can weaken boundary-specific cues or propagate local corrections into otherwise stable region interiors.

Our proposed approach, AURASeg, addresses this problem with an explicit \emph{reconstruct-then-correct} approach. First, the Attention Progressive Upsampling Decoder (APUD) progressively combines deep semantic features with shallower spatial features to recover the main drivable-area representation. The Residual Boundary Refinement Module (RBRM) then operates on the reconstructed feature map, estimates boundary-sensitive residual features, and injects only a learned gated correction. ASPPLite provides compact multi-scale context at the bottleneck but is used as a supporting context module to the entire architecture. Our contributions are summarized as follows:
\begin{enumerate}
    \item an \textit{Attention Progressive Upsampling Decoder} (APUD) that aligns semantic and detail features in a shared 128-channel space and combines channel-attended semantic responses with spatially attended skip features through selective multiplicative fusion;
    \item a \textit{Residual Boundary Refinement Module} (RBRM) that derives Sobel-informed boundary features with a compact encoder--decoder and injects them as a learned residual correction after region reconstruction; and
    \item a boundary-centric evaluation of the above two sequential design choices, together with controlled APUD and RBRM ablations, qualitative analysis, and common-setting hardware efficiency evaluation.
\end{enumerate}

%% file: sec/2_formatting.tex
\section{Related Work}
\label{sec:related_work}

\noindent\textbf{Efficient encoder--decoder segmentation.}
Semantic segmentation for autonomous navigation must preserve high-level scene understanding without discarding the spatial detail needed to separate traversable ground from nearby obstacles. Ghost-UNet~\cite{c7} uses an asymmetric encoder--decoder for semantic segmentation from scratch, while residual connections~\cite{c10} remain a standard mechanism for maintaining information flow and stabilizing deep networks. Real-time designs pursue stronger accuracy--efficiency trade-offs through bilateral or multi-branch processing: BiSeNet~\cite{c4} separates spatial and contextual pathways, TwinLiteNet~\cite{c13} targets compact joint drivable-area and lane segmentation, and PIDNet~\cite{c28} maintains dedicated detail, context, and boundary branches. Other efficient road-scene architectures explore complementary strategies for practical deployment. SwiftNet combines a lightweight pretrained encoder with lateral upsampling for road-driving segmentation \cite{swiftnet2019}, DDRNet maintains parallel high- and low-resolution representations with repeated bilateral fusion \cite{ddrnet2021}, and STDC integrates detail learning into an efficient single-stream architecture \cite{stdc2021}. YOLOP approaches drivable-area perception from a multi-task perspective by jointly predicting objects, lanes, and drivable regions, including embedded-device evaluation \cite{yolop2022}. These models emphasize efficient scene parsing, while our approach specifically examines how decoder feature recovery and subsequent boundary refinement affect free-space contour localization.

\noindent\textbf{Multi-scale context and attention.}
Parallel dilated convolutions provide a practical way to enlarge the receptive field while retaining a dense feature map. S2-FPN~\cite{c12} introduces scale-aware strip-attention connections, while depth-guided free-space segmentation~\cite{c11} shows how geometric cues can complement appearance information in robotic scenes. Attention provides another mechanism for selective feature recovery: Squeeze-and-Excitation (SE)~\cite{c19} reweights channels, CBAM~\cite{c20} combines channel and spatial attention, and feature-refinement modules~\cite{c14} aggregate multi-stage semantic features and non-local context. Attention has also been incorporated directly into decoder upsampling. PAN introduces a Global Attention Upsample module in which high-level global information guides the selection of low-level localization features \cite{pan2018}, while AUNet develops an attention-guided dense-upsampling block that combines dense upsampling, high/low-level feature fusion, and channel attention \cite{aunet2020}. These methods motivate selective feature recovery during decoding.

\noindent\textbf{Boundary-aware segmentation.}
Boundary-aware segmentation methods explicitly model edges or detail features to sharpen thin structures and object transitions~\cite{c15,c30}. FBRNet~\cite{c3} couples cross-scale feature fusion with a dedicated border-refinement pathway, whereas BASeg~\cite{c30} uses a boundary prior and boundary-guided context aggregation. Street-floor segmentation~\cite{c16} addresses traversability-related surface understanding for wheeled robots. RGB-D approaches have also explored uncertainty-aware free-driving-region detection~\cite{c18} and self-supervised drivable-area and road-anomaly segmentation~\cite{c21}. These approaches motivate explicit contour modeling, but boundary cues may be introduced during feature extraction, fused throughout the network, or applied as a later refinement stage.

\noindent\textbf{Positioning of AURASeg.}
AURASeg follows the last of these strategies. APUD first reconstructs the drivable-area representation from semantic and spatial features, and RBRM then estimates a boundary-sensitive residual from the completed decoder representation. The stepwise analysis isolates the contribution of each component in the proposed architecture and tests whether
it improves boundary quality relative to simpler decoder alternatives as evaluated in \cref{sec:results}.

%% file: sec/3_finalcopy.tex
\section{Proposed Method}
\label{sec:method}

\begin{figure*}[t]
  \centering
  \includegraphics[width=\textwidth,trim=0 0 0 4mm,clip]{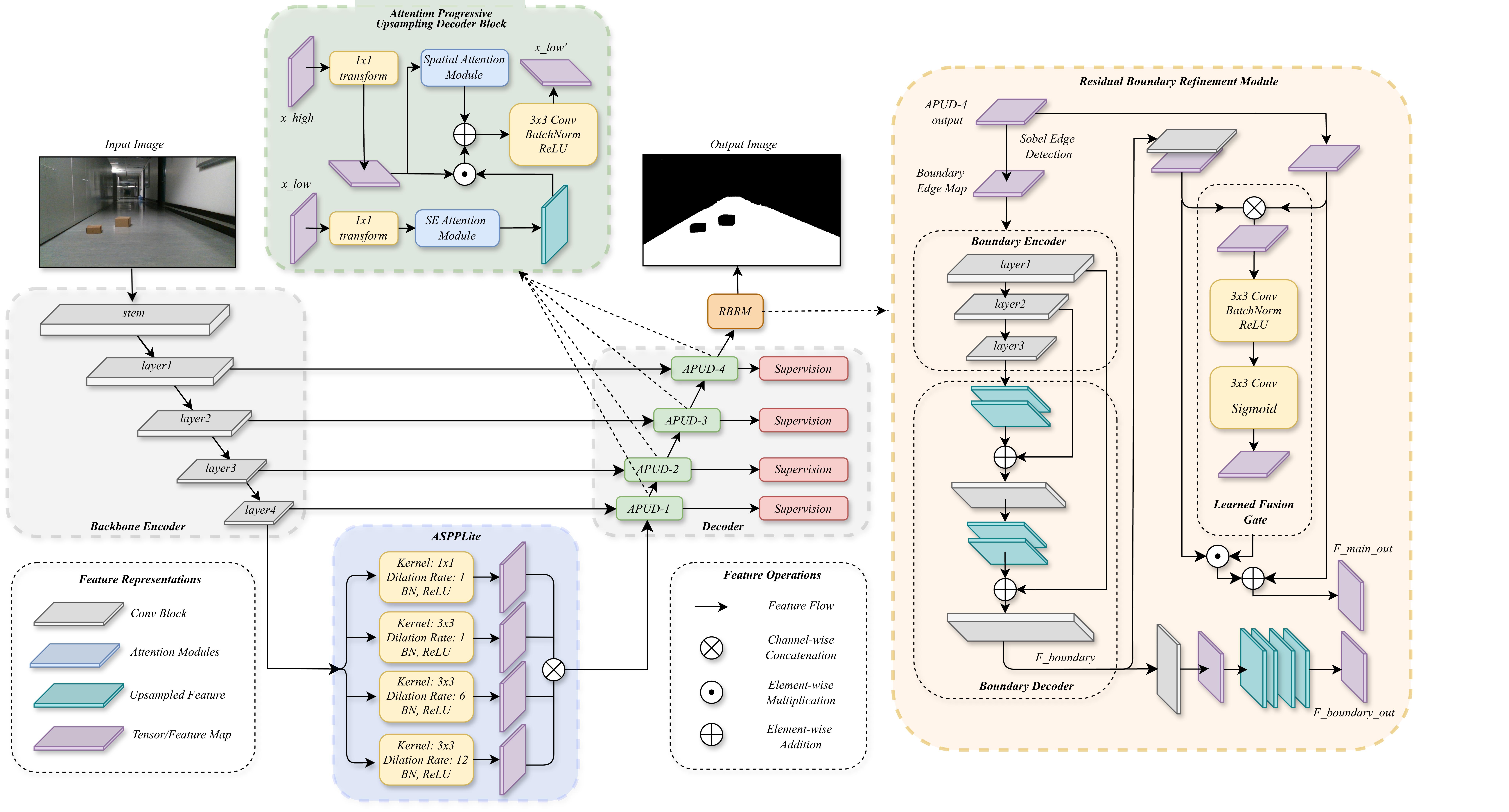}
  \caption{AURASeg architecture. A ResNet-18 encoder provides hierarchical features to ASPPLite and four APUD stages. APUD reconstructs the drivable-region representation in a shared 128-channel space; RBRM then estimates and injects a boundary-sensitive residual correction. Auxiliary APUD and boundary heads are used only for training.}
  \label{fig:2}
\end{figure*}

\subsection{Architecture Overview}
As shown in \cref{fig:2}, AURASeg is designed as a sequential \emph{reconstruct-then-correct} segmentation model. A ResNet-18 encoder extracts hierarchical features at progressively lower spatial resolutions. ASPPLite supplies multi-scale bottleneck context, after which four Attention Progressive Upsampling Decoder (APUD) blocks reconstruct a high-resolution drivable-area representation. The Residual Boundary Refinement Module (RBRM) is deliberately applied only after this reconstruction. It estimates a boundary-specific residual and uses a learned gate to control where that residual modifies the region feature. This ordering separates region recovery from contour correction rather than mixing boundary cues throughout the decoder.

\subsection{Encoder and ASPPLite Context}
The encoder is an ImageNet-pretrained ResNet-18~\cite{c10}. Its skip features have 64, 64, 128, and 256 channels, while the deepest bottleneck feature has 512 channels. ASPPLite processes this 512-channel tensor with four parallel branches: one $1\times1$ convolution and three $3\times3$ atrous convolutions with dilation rates $1$, $6$, and $12$. Each branch produces 64 channels and uses BatchNorm followed by SiLU. The four outputs are concatenated to 256 channels, projected to the 128-channel decoder width by a $1\times1$ Conv--BN--SiLU block, and regularized with dropout $p{=}0.1$. ASPPLite is used as a compact supporting context module; the central architectural contribution lies in the APUD reconstruction stage and the subsequent RBRM correction stage.

\subsection{Attention Progressive Upsampling Decoder}
Each APUD block receives a deeper low-resolution semantic tensor $x_{\mathrm{low}}$ and a shallower high-resolution skip tensor $x_{\mathrm{high}}$. Both are projected to $C{=}128$ channels using $1\times1$ Conv--BN--SiLU layers. Channel attention is applied to the semantic branch using Squeeze-and-Excitation (SE)~\cite{c19}; the result is bilinearly upsampled to the spatial resolution of the skip feature. In parallel, spatial attention~\cite{c20} is computed from channel-wise average and max responses of the projected skip tensor using a $7\times7$ convolution and sigmoid mask.

The default APUD fusion multiplies the upsampled semantic response with the projected skip feature, then adds the spatially attended skip response:
\begin{equation}
L=T_{\ell}(x_{\mathrm{low}}), \qquad H=T_h(x_{\mathrm{high}}),
\end{equation}
\begin{equation}
L'=\mathcal{U}(\mathrm{SE}(L)), \qquad F=L'\odot H, \qquad S=\mathrm{SA}(H),
\end{equation}
\begin{equation}
y=R(F+S),
\end{equation}
where $T_{\ell}$ and $T_h$ are the $1\times1$ projections, $\mathcal{U}$ denotes bilinear interpolation, $\odot$ is element-wise multiplication, and $R$ is a $3\times3$ Conv--BN--SiLU refinement block. The four APUD stages retain the same 128-channel width while progressively moving from the bottleneck toward the $H/4$ representation. Keeping a common decoder width makes the semantic/detail interaction identical at each scale and avoids introducing stage-specific fusion rules. Each stage also has an auxiliary $3\times3$--$1\times1$ segmentation head (128$\rightarrow$32$\rightarrow$2 channels) for deep supervision during training. These auxiliary logits are resized to the input resolution and are discarded during inference, so they affect optimization but not the deployed prediction path.

\subsection{Residual Boundary Refinement Module}
RBRM takes the final APUD feature $\mathbf{F}\in\mathbb{R}^{128\times H/4\times W/4}$. A $1\times1$ projection first reduces it to 64 channels. Fixed, depth-wise Sobel-$x$ and Sobel-$y$ kernels are then applied independently to every channel; the two responses are concatenated and fused with a learned $3\times3$ Conv--BN--SiLU layer to obtain an edge-sensitive tensor $\mathbf{E}\in\mathbb{R}^{64\times H/4\times W/4}$.

A three-stage boundary encoder increases channels $64\rightarrow128\rightarrow256\rightarrow512$ while downsampling by stride-2 convolutions. A symmetric decoder bilinearly upsamples and concatenates the corresponding encoder skips, finally reconstructing a 64-channel boundary representation $\mathbf{B}$. A learned projection maps $\mathbf{B}$ back to 128 channels, and a gate is estimated from the concatenation of the reconstructed region feature and projected boundary feature:
\begin{equation}
\mathbf{P}=\phi(\mathbf{B}), \qquad
\mathbf{G}=\sigma\!\left(\mathrm{BN}\!\left(\mathrm{Conv}_{1\times1}([\mathbf{F},\mathbf{P}])\right)\right),
\end{equation}
\begin{equation}
\mathbf{F}_{\mathrm{ref}}=\mathbf{F}+\mathbf{G}\odot\mathbf{P}.
\end{equation}

Applying boundary refinement after the main decoder representation is intentional. The progressive decoder is first allowed to establish a stable semantic estimate of the drivable region using information from multiple feature scales. Boundary-sensitive processing is then applied to this already formed representation, rather than being repeatedly introduced at every decoder stage. This separation reduces the chance that local edge responses alter otherwise reliable interior predictions. Moreover, the refinement branch does not directly replace the decoder feature. Instead, the learned gate determines the spatial locations and magnitude at which the boundary-derived residual contributes to the final representation. Consequently, the module can concentrate its additional capacity near uncertain transitions while preserving features in regions that already exhibit consistent foreground evidence. This design provides a direct connection between the architectural objective and the evaluation protocol, where improvements in contour agreement are considered together with preservation of region-level accuracy.

Thus, RBRM learns a residual correction rather than replacing the region representation. The residual form is important to the reconstruct-then-correct design: interior regions can retain the APUD estimate while boundary-sensitive locations receive an additional learned update. A separate $1\times1$ boundary head on $\mathbf{B}$ produces one boundary logit map for auxiliary supervision. The final segmentation head applies a $3\times3$ Conv--BN--SiLU block (128$\rightarrow$64), dropout $p{=}0.1$, and a $1\times1$ classifier to two classes before bilinear upsampling to the input size.

\begin{table}[!t]
\centering
\caption{AURASeg training configuration. The implementation uses deterministic seed 42, differential encoder/decoder learning rates, and gradient accumulation to maintain effective batch size 8.}
\label{config}
\footnotesize
\setlength{\tabcolsep}{3.5pt}
\renewcommand{\arraystretch}{1.07}
\begin{tabularx}{\columnwidth}{@{} >{\raggedright\arraybackslash}p{0.43\columnwidth} X @{}}
\toprule
\textbf{Setting} & \textbf{Value} \\
\midrule
Input / classes & $384\times640$ / 2 \\
Encoder & ResNet-18, ImageNet pretrained \\
APUD width & 128 channels \\
Optimizer & AdamW ($\beta_1{=}0.9,\ \beta_2{=}0.999$) \\
Weight decay & 0.01 \\
LR (encoder / decoder) & $10^{-4}$ / $10^{-3}$ \\
Scheduler & CosineAnnealingLR, $\eta_{\min}{=}10^{-6}$ \\
Epochs / early stop & 50 / patience 10, $\Delta{=}10^{-4}$ \\
Micro-batch / accumulation & 4 / 2 (effective batch 8) \\
Validation batch & 4 \\
Mixed precision & AMP FP16 \\
Training GPU & RTX 5060 GPU \\
Workers / pin memory & 4 / enabled \\
\addlinespace[1pt]
Main region loss & $0.5\,\mathcal{L}_{\mathrm{Focal}}+0.5\,\mathcal{L}_{\mathrm{Dice}}$ \\
Focal / Dice parameters & $\alpha{=}0.25$, $\gamma{=}2$ / smooth $=1$ \\
Boundary supervision & $0.2\,\mathcal{L}_{\mathrm{BCE}}$ \\
Deep supervision & $0.1\sum_{k=1}^{4}(\mathcal{L}^{(k)}_{\mathrm{Focal}}+\mathcal{L}^{(k)}_{\mathrm{Dice}})$ \\
\bottomrule
\end{tabularx}
\end{table}

\begin{table*}[t]
\centering
\caption{Datasets used for drivable-area segmentation. MIX denotes the union of the held-out Gazebo and GMRPD test partitions used for the MIX test benchmark; it is not an additional dataset.}
\label{dataset}
\setlength{\tabcolsep}{6pt}
\renewcommand{\arraystretch}{1.10}
\begin{tabular}{l l c c c c}
\toprule
\textbf{Dataset} & \textbf{Environment} & \textbf{Train} & \textbf{Val.} & \textbf{Test} & \textbf{Total} \\
\midrule
Gazebo (ours) & Indoor laboratory corridors (simulation) & 2483 & 294 & 420 & 3197 \\
GMRPD~\cite{c21} & Outdoor sidewalks, plazas, and squares & 3000 & 360 & 536 & 3896 \\
\mbox{CARL-D}~\cite{c22} & On-road scenes from a driver-view camera & 8372 & 1046 & 1046 & 10464 \\
\bottomrule
\end{tabular}
\end{table*}

\subsection{Objective and Optimization}
AURASeg is trained with three complementary supervision terms. For target mask $Y$, main logits $\hat{Y}$, auxiliary APUD logits $\hat{Y}^{(k)}$, and RBRM boundary logits $\hat{B}$, the total loss is
\begin{equation}
\label{eq:total_loss}
\mathcal{L}=\mathcal{L}_{\mathrm{main}}+0.2\,\mathcal{L}_{\mathrm{BCE}}(\hat{B},B)+0.1\sum_{k=1}^{4}\mathcal{L}^{(k)}_{\mathrm{aux}},
\end{equation}
where
\begin{equation}
\mathcal{L}_{\mathrm{main}}=0.5\,\mathcal{L}_{\mathrm{Focal}}(\hat{Y},Y)+0.5\,\mathcal{L}_{\mathrm{Dice}}(\hat{Y},Y),
\end{equation}
\begin{equation}
\mathcal{L}^{(k)}_{\mathrm{aux}}=\mathcal{L}_{\mathrm{Focal}}(\hat{Y}^{(k)},Y)+\mathcal{L}_{\mathrm{Dice}}(\hat{Y}^{(k)},Y).
\end{equation}
The focal term is computed from per-pixel cross entropy as $\alpha(1-p_t)^\gamma\mathcal{L}_{\mathrm{CE}}$ with $\alpha{=}0.25$ and $\gamma{=}2$. Dice loss applies softmax probabilities, one-hot targets, and averages the class-wise Dice score with smoothing 1.0. For boundary supervision, the binary target is generated directly from the semantic mask using a $3\times3$ morphological gradient,
\begin{equation}
B=\mathrm{dilate}_{3\times3}(Y)-\mathrm{erode}_{3\times3}(Y),
\end{equation}
and optimized with unweighted \texttt{BCEWithLogitsLoss}. Nearest-neighbor interpolation is used whenever a discrete mask must be resized.

Optimization follows the configuration in \cref{config}. Training uses horizontal flipping ($p{=}0.5$), shift/scale/rotation ($\pm0.1/\pm0.1/\pm15^\circ$, $p{=}0.5$), brightness/contrast perturbation ($\pm0.2$, $p{=}0.3$), and Gaussian noise with variance range $(10,50)$ ($p{=}0.2$), followed by ImageNet normalization. Model selection uses validation mIoU; the best checkpoint must exceed the previous best by $10^{-4}$, otherwise the patience counter is incremented. The final test evaluation always reloads the best validation checkpoint. This keeps model selection separate from the reported test measurements and ensures that every final result corresponds to the same validation-based selection rule.

%% file: sec/4_results.tex
\section{Results}
\label{sec:results}

\subsection{Datasets and Evaluation Protocol}
As summarized in \cref{dataset}, the evaluation spans three complementary settings: indoor simulation (Gazebo), outdoor ground-robot imagery from the Ground Mobile Robot Perception Dataset (GMRPD~\cite{c21}), and road scenes from \mbox{CARL-D}~\cite{c22}. Only RGB images and binary drivable-area masks are used. MIX denotes the fixed union of the held-out Gazebo test partition (420 images) and the GMRPD test partition (536 images). The corresponding training partitions are used for model training, while the validation partitions are used only for checkpoint selection. The purpose of this combined test benchmark is to evaluate the same trained model across simulated indoor structure and outdoor pedestrian-scale robot scenes rather than treating them as isolated operating conditions.

For \mbox{CARL-D}, 10,464 image--mask pairs are split into 8,372/1,046/1,046 train/validation/test samples. RGB masks are decoded before resizing: road $(17,163,74)$ and observed black fallback pixels $(0,0,0)$ map to class 1, while background $(15,16,65)$ maps to class 0. Unexpected RGB values raise an error, and masks are resized only with nearest-neighbor interpolation. All models are evaluated at $384\times640$, so differences in the reported metrics are not attributable to test-time resolution.

Region metrics are computed from globally accumulated foreground pixels. We report drivable-class IoU and F1 in the compact comparison tables, with precision and recall retained in the evaluation outputs. Boundary metrics are computed per image and macro-averaged. A $3\times3$ morphological gradient extracts predicted and target boundaries, and both are dilated by $k{=}2$ iterations before boundary IoU and F1 are computed. The region metrics therefore measure overall foreground agreement, whereas BIoU/BF1 place more emphasis on whether the predicted contour is spatially aligned with the target boundary.

The general comparison includes FCN-R50~\cite{c23}, PSPNet-R50~\cite{c24}, UPerNet-R50~\cite{c25}, SegFormer-B2~\cite{c26}, Mask2Former~\cite{c27}, and PIDNet-L~\cite{c28}. FBRNet~\cite{c3} and BASeg~\cite{c30} provide explicit boundary-focused comparisons. All comparison models use the same dataset partitions, input resolution, preprocessing, augmentation framework, checkpoint-selection criterion, and evaluation implementation. Common optimization settings are used where they are compatible with the corresponding architectures.

\begin{table*}[!t]
\centering
\caption{Stepwise AURASeg ablation on the MIX test benchmark with a ResNet-18 backbone. The cumulative design shows how context aggregation, attention-guided reconstruction, and residual boundary correction change model size and boundary quality.}
\label{tab:ablation_resnet18}
\footnotesize
\setlength{\tabcolsep}{5.0pt}
\renewcommand{\arraystretch}{1.08}
\begin{tabular}{lcccccc}
\toprule
\textbf{Variant} & \textbf{ASPPLite} & \textbf{APUD} & \textbf{RBRM} & \textbf{Params. (M)} & \textbf{Acc.$^{\dagger}$} & \textbf{BF1} \\
\midrule
V1: Base (ResNet-18) & -- & -- & -- & 14.61 & 0.9928 & 0.7804 \\
V2: V1 + ASPPLite & \cmark & -- & -- & 17.98 & 0.9928 & 0.7931 \\
V3: V2 + APUD & \cmark & \cmark & -- & 18.21 & 0.9943 & 0.8224 \\
V4: AURASeg & \cmark & \cmark & \cmark & \textbf{20.86} & \textbf{0.9946} & \textbf{0.8905} \\
\bottomrule
\end{tabular}
\vspace{1mm}
\begin{flushleft}
\footnotesize
\centering
$^{\dagger}$ Pixel accuracy computed over all pixels.
\end{flushleft}
\end{table*}

\begin{figure*}[!t]
  \centering
  \includegraphics[width=\textwidth]{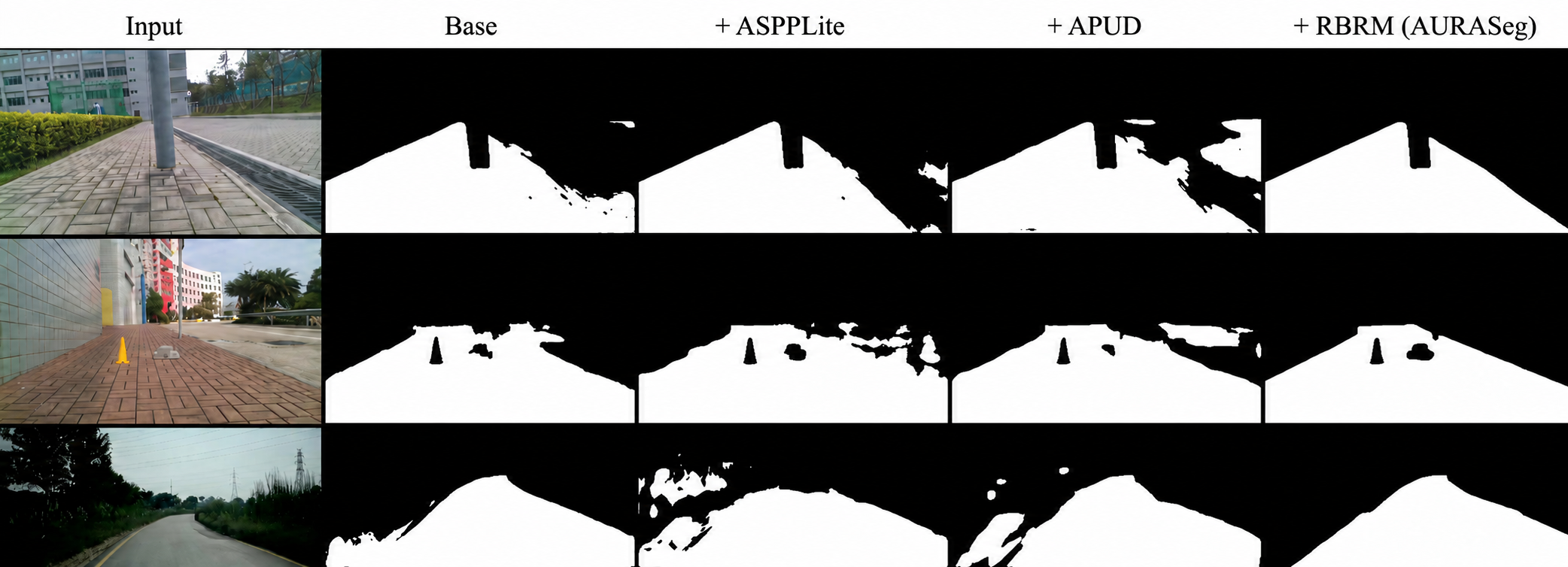}
  \caption{Qualitative progression through the stepwise ablation: ResNet-18 baseline, ASPPLite context, APUD, and RBRM correction.}
  \label{fig:ablation_study}
\end{figure*}

\subsection{Main Quantitative Comparison}
Tables~\ref{metrics} and~\ref{deploy_metrics}  summarize two complementary aspects of the final system: segmentation quality and deployment cost. The first separates region agreement from contour agreement, while the second provides the corresponding embedded-efficiency reference for the same 128-channel AURASeg design.

\begin{table*}[!t]
\centering
\small
\caption{Main segmentation comparison. Panel (a) covers general segmentation baselines; panel (b) isolates the boundary-focused external comparison. IoU/F1 measure foreground drivable region, whereas BIoU/BF1 measure contour agreement under the common $k{=}2$ protocol.}
\label{metrics}
\renewcommand{\arraystretch}{1.10}
\begin{tabular*}{\textwidth}{@{\extracolsep{\fill}}lcccccccc@{}}
\toprule
\multicolumn{9}{c}{\textbf{(a) General segmentation baselines}}\\
\midrule
\multirow{2}{*}{\textbf{Model}} &
\multicolumn{4}{c}{\textbf{MIX test}} &
\multicolumn{4}{c}{\textbf{\mbox{CARL-D} test}} \\
\cmidrule(lr){2-5}\cmidrule(lr){6-9}
& \textbf{IoU} & \textbf{F1} & \textbf{BIoU} & \textbf{BF1}
& \textbf{IoU} & \textbf{F1} & \textbf{BIoU} & \textbf{BF1} \\
\midrule
FCN-R50~\cite{c23}        & 0.9857 & 0.9928 & 0.6502 & 0.7789 & 0.9575 & 0.9783 & 0.5269 & 0.6691 \\
PSPNet-R50~\cite{c24}     & 0.9870 & 0.9935 & 0.7639 & 0.8589 & 0.9579 & 0.9785 & 0.5265 & 0.6683 \\
UPerNet-R50~\cite{c25}    & 0.9879 & 0.9939 & 0.7863 & 0.8738 & 0.9567 & 0.9779 & 0.5299 & 0.6721 \\
SegFormer-B2~\cite{c26}   & 0.9885 & 0.9942 & 0.7763 & 0.8683 & 0.9574 & 0.9782 & 0.5263 & 0.6671 \\
Mask2Former~\cite{c27}    & 0.9881 & 0.9940 & 0.7740 & 0.8661 & 0.9450 & 0.9717 & 0.4491 & 0.5999 \\
PIDNet-L~\cite{c28}       & 0.9835 & 0.9917 & 0.6334 & 0.7656 & \textbf{0.9584} & \textbf{0.9788} & 0.5391 & 0.6791 \\
\midrule
\textbf{AURASeg (Ours)}   & \textbf{0.9897} & \textbf{0.9948} & \textbf{0.8124} & \textbf{0.8905} & 0.9547 & 0.9768 & \textbf{0.5424} & \textbf{0.6811} \\
\bottomrule
\end{tabular*}

\vspace{4pt}
\begin{tabular*}{\textwidth}{@{\extracolsep{\fill}}lcccccccc@{}}
\toprule
\multicolumn{9}{c}{\textbf{(b) Boundary-focused comparison}}\\
\midrule
\multirow{2}{*}{\textbf{Model}} &
\multicolumn{4}{c}{\textbf{MIX test}} &
\multicolumn{4}{c}{\textbf{\mbox{CARL-D} test}} \\
\cmidrule(lr){2-5}\cmidrule(lr){6-9}
& \textbf{IoU} & \textbf{F1} & \textbf{BIoU} & \textbf{BF1}
& \textbf{IoU} & \textbf{F1} & \textbf{BIoU} & \textbf{BF1} \\
\midrule
FBRNet~\cite{c3}         & 0.9860 & 0.9929 & 0.6669 & 0.7922 & 0.9565 & 0.9778 & 0.5320 & 0.6737 \\
BASeg~\cite{c30}         & 0.9875 & 0.9937 & 0.7660 & 0.8609 & \textbf{0.9592} & \textbf{0.9791} & 0.5294 & 0.6721 \\
\textbf{AURASeg (Ours)}  & \textbf{0.9897} & \textbf{0.9948} & \textbf{0.8124} & \textbf{0.8905} & 0.9547 & 0.9768 & \textbf{0.5424} & \textbf{0.6811} \\
\bottomrule
\end{tabular*}
\end{table*}

\begin{table*}[!t]
\centering
\small
\caption{Hardware deployment comparison at $384\times640$ and batch size 1 on a Jetson Nano 4GB device.}
\label{deploy_metrics}
\renewcommand{\arraystretch}{1.10}

\begin{tabular*}{\textwidth}{@{\extracolsep{\fill}}lccccc@{}}
\toprule
\textbf{Model} &
\textbf{Params. (M)} &
\textbf{GFLOPs} &
\textbf{Peak mem. (MB)} &
\textbf{Latency (ms)} &
\textbf{FPS} \\
\midrule

FCN-R50~\cite{c23}
    & 35.31 & 139.25 & 265.5 & 843.2 & 1.19 \\

PSPNet-R50~\cite{c24}
    & 24.30 & 11.09 & 125.1 & 60.6 & 16.50 \\

UPerNet-R50~\cite{c25}
    & 37.28 & 71.93 & 478.2 & 523.4 & 1.91 \\

SegFormer-B2~\cite{c26}
    & 24.72 & 19.86 & 285.4 & 418.6 & 2.39 \\

Mask2Former~\cite{c27}
    & 45.96 & 36.81 & 202.1 & 361.3 & 2.77 \\

PIDNet-L~\cite{c28}
    & 36.94 & 32.35 & 192.3 & 210.3 & 4.76 \\

FBRNet~\cite{c3}
    & 17.16 & 16.29 & 130.9 & 99.6 & 10.04 \\

BASeg-R101~\cite{c30}
    & 78.45 & 306.77 & 349.2 & 950.1 & 1.05 \\

\textbf{AURASeg (Ours)}
    & \textbf{20.86} & 72.70 & 261.5 & 321.0 & 3.12 \\

\bottomrule
\end{tabular*}

\vspace{1mm}

\end{table*}

The two panels in \cref{metrics} answer different questions. Panel (a) places AURASeg among established segmentation architectures and therefore indicates whether stronger contour localization is obtained without losing competitive region segmentation. Panel (b) narrows the comparison to methods that explicitly emphasize boundary refinement, providing a more direct test of the role targeted by RBRM.

\paragraph{General segmentation baselines.}
On MIX, AURASeg gives the strongest result in every reported column: IoU $0.9897$, F1 $0.9948$, BIoU $0.8124$, and BF1 $0.8905$. The region scores are already close across the strongest models, so the more informative separation appears at the contour. Relative to UPerNet-R50, the next-best MIX boundary entry, AURASeg improves BIoU from $0.7863$ to $0.8124$ and BF1 from $0.8738$ to $0.8905$. This pattern is consistent with the intended behavior of the architecture: APUD preserves the broad free-space region, while the additional refinement capacity is expressed more clearly in the boundary metrics than in already-saturated region overlap.

On \mbox{CARL-D}, the ranking exposes a different region--boundary trade-off. PIDNet-L gives the highest region IoU/F1 at $0.9584/0.9788$, while AURASeg records $0.9547/0.9768$. In contrast, AURASeg gives the strongest boundary result in the general comparison, with BIoU $0.5424$ and BF1 $0.6811$, compared with $0.5391/0.6791$ for PIDNet-L. The margins are modest, but the ordering is consistent across both boundary measures. Thus, the road-scene result should not be interpreted as uniform dominance: AURASeg trades a small amount of region overlap for the best contour agreement among the evaluated general baselines.

\paragraph{Boundary-focused baselines.}
Panel (b) tests that interpretation against FBRNet and BASeg. On MIX, AURASeg again leads all four reported metrics; its BF1 of $0.8905$ is higher than BASeg's $0.8609$ and FBRNet's $0.7922$, while its IoU/F1 also remain highest. On \mbox{CARL-D}, BASeg gives the strongest region IoU/F1 ($0.9592/0.9791$), followed by FBRNet ($0.9565/0.9778$), whereas AURASeg gives the highest BIoU/BF1 ($0.5424/0.6811$). The same separation therefore appears in both the general and boundary-focused comparisons: AURASeg is most competitive when contour localization is emphasized, while some alternatives retain a small advantage in road-scene region overlap.

Taken together, the two evaluation settings reveal a consistent characteristic of the proposed model rather than a uniform advantage on every metric. On MIX, where several models already achieve very high foreground overlap, the differences between strong methods become substantially more visible in the boundary measures. AURASeg improves both region and contour scores in this setting, with the larger separation appearing in boundary IoU and boundary F1. CARL-D presents a more challenging trade-off: PIDNet-L and BASeg obtain slightly higher region overlap in their respective comparison groups, whereas AURASeg produces the highest boundary scores in both panels. This distinction is important because a high region IoU can remain largely unchanged when errors occupy only a narrow band around the drivable-area contour. Such errors contribute relatively few pixels to the complete foreground region but directly affect the spatial placement of the free-space boundary. The results therefore suggest that the contribution of AURASeg is expressed most consistently through improved contour localization while maintaining region predictions within the competitive range of the evaluated models. This behavior is also reflected by the subsequent component studies, where changes to the proposed decoder and refinement stages produce clearer variations in the boundary metrics than in overall region overlap.

\subsection{Efficiency and Deployment}
The 128-channel APUD configuration keeps the complete four-stage decoder and full RBRM while limiting the model to 20.86M parameters and approximately 72.70 GFLOPs at $384\times640$. Training is performed on the RTX 5060 GPU reported in \cref{config}; the deployment comparison in \cref{deploy_metrics} is separated because training hardware and embedded inference hardware serve different purposes.

\Cref{deploy_metrics} compares direct Jetson Nano measurements for the models discussed. AURASeg uses 20.86M parameters, which is smaller than every listed general segmentation baseline, while FBRNet remains lighter among the boundary-focused models. Its 72.70 GFLOPs place it near UPerNet-R50 in arithmetic cost, and the 321.0 ms latency (3.12 FPS) lies between the faster PSPNet-R50, FBRNet, and PIDNet-L entries and the slower Mask2Former, SegFormer-B2, UPerNet-R50, FCN-R50, and BASeg-R101 entries. The result is consistent with the architecture: the 128-channel APUD width controls model size, while the multi-stage boundary encoder--decoder in RBRM still carries a measurable deployment cost. 

\subsection{Ablation and Boundary Sensitivity}
The cumulative ablation in \cref{tab:ablation_resnet18} establishes how the main stages change boundary quality before the finer mechanism test is considered. Adding ASPPLite raises BF1 from $0.7804$ to $0.7931$ without changing accuracy, indicating a modest benefit from additional context alone. Introducing APUD increases BF1 to $0.8224$ and accuracy to $0.9943$, and the subsequent RBRM raises BF1 to $0.8905$ and accuracy to $0.9946$. The progression therefore supports the proposed division of labor: APUD reconstructs the semantic/detail representation, and the subsequent RBRM contributes the final boundary-focused correction after that representation has been formed.

The controlled APUD comparison in \cref{tab:diagnostics} isolates two decoder choices without turning the diagnostic into a search over many nearby variants. Replacing multiplicative fusion with addition causes the clearest degradation: BIoU falls from $0.8124$ to $0.7849$ and BF1 from $0.8905$ to $0.8730$. Removing channel and spatial attention also lowers boundary quality, to $0.8046$ BIoU and $0.8858$ BF1, while region IoU changes only slightly. The disproportionate change in boundary metrics supports the interpretation that the APUD fusion and attention operations primarily improve local spatial recovery rather than merely increasing foreground coverage.

The tolerance portion of \cref{tab:diagnostics} provides context for the absolute boundary scores. Increasing $k$ relaxes the localization requirement and therefore raises BIoU/BF1, whereas $k{=}1$ is stricter. The main experiments use $k{=}2$ for every model; the $k{=}1$ and $k{=}3$ values are reported only to show the sensitivity of the same predictions to the permitted contour displacement.

\begin{table}[!t]
\centering
\footnotesize
\caption{Compact diagnostic analysis on MIX. (a) Controlled APUD alternatives; (b) sensitivity of the same AURASeg predictions to boundary tolerance $k$.}
\label{tab:diagnostics}
\setlength{\tabcolsep}{4.3pt}
\renewcommand{\arraystretch}{1.06}
\begin{tabular}{lccc}
\toprule
\multicolumn{4}{c}{\textbf{(a) APUD mechanism ablation}}\\
\midrule
\textbf{Variant} & \textbf{IoU} & \textbf{BIoU} & \textbf{BF1} \\
\midrule
Additive fusion & 0.9887 & 0.7849 & 0.8730 \\
No attention    & 0.9892 & 0.8046 & 0.8858 \\
\textbf{AURASeg (Ours)} & \textbf{0.9897} & \textbf{0.8124} & \textbf{0.8905} \\
\midrule
\multicolumn{4}{c}{\textbf{(b) Boundary tolerance sensitivity}}\\
\midrule
\textbf{$k$} & \multicolumn{1}{c}{\textbf{BIoU}} & \multicolumn{1}{c}{\textbf{BF1}} & \\
\midrule
1 & 0.7696 & 0.8621 & \\
\textbf{2} & \textbf{0.8124} & \textbf{0.8905} & \\
3 & 0.8369 & 0.9056 & \\
\bottomrule
\end{tabular}
\end{table}

\begin{figure}[t]
  \centering
  \includegraphics[width=\columnwidth]{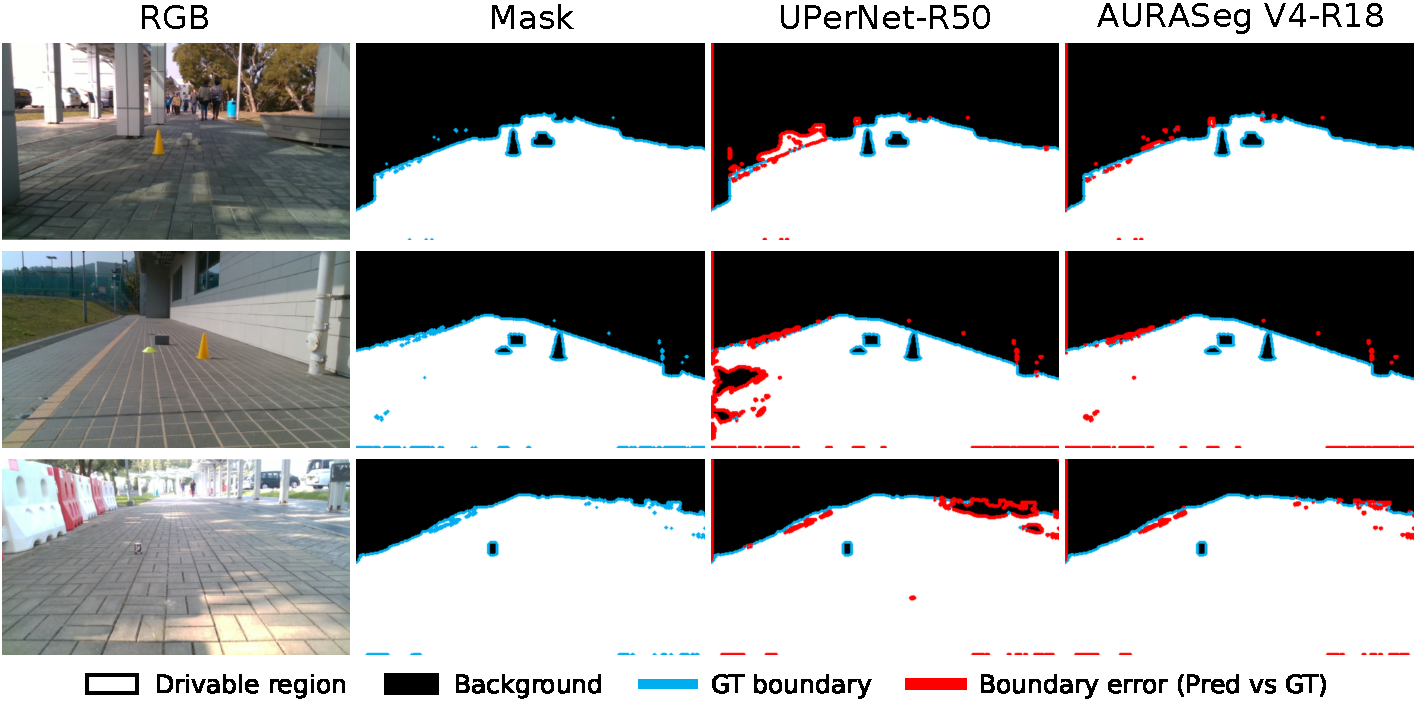}
  \caption{Qualitative boundary comparison between UPerNet-R50 and AURASeg. Red denotes boundary errors relative to the ground-truth boundary (blue).}
  \label{fig:qualitative}
\end{figure}

\subsection{Qualitative Analysis}
The qualitative examples complement the aggregate metrics by showing how the refinement appears spatially. In \cref{fig:ablation_study}, the baseline and context-only stages contain fragmented or displaced free-space edges; APUD recovers a more coherent region, and the final RBRM stage further regularizes the contour around poles, curbs, and scene transitions. \Cref{fig:qualitative} provides a direct boundary view: around thin obstacles and cluttered transitions, AURASeg prediction more closely follows the reference contour in several locations where a small inward or outward displacement would have little effect on total foreground overlap. These examples help in explaining why relatively small changes in region IoU can correspond to clearer changes in BIoU/BF1.

\subsection{Limitations}
The results also identify clear limitations. First, the strongest boundary ranking on \mbox{CARL-D} does not coincide with the strongest region ranking: PIDNet-L leads the general baselines in IoU/F1 and BASeg leads the boundary-focused group in IoU/F1. AURASeg therefore emphasizes contour agreement rather than uniformly dominating region overlap. Second, BIoU/BF1 depend on the spatial tolerance used to match contours. We fix $k{=}2$ for every main comparison and report the sensitivity analysis to make this dependency explicit, but a single tolerance cannot represent every downstream navigation requirement. Finally, AURASeg operates frame-by-frame and does not model temporal consistency, so boundary estimates may still fluctuate in video. 

%% file: sec/5_conclusion.tex
\section{Conclusion}
\label{sec:conclusion}

AURASeg demonstrates that boundary-aware refinement can improve drivable-area segmentation quality while retaining strong region-level predictions. Across the complete evaluation, this behavior remains consistent in the stepwise ablation, the controlled decoder study, and the comparisons with both general-purpose and boundary-focused segmentation models. The experimental results show a clear improvement in boundary-oriented metrics as the proposed components are introduced, while overall region overlap and prediction consistency remain strong across the evaluated indoor, outdoor, and road-scene datasets. This indicates that the attention-guided upsampling stage and the subsequent residual boundary refinement are complementary: the former improves the recovery of spatial detail during decoding, while the latter concentrates additional refinement on contour-sensitive regions before the final prediction. The resulting model therefore achieves a more balanced trade-off between region accuracy and boundary localization rather than optimizing one at the expense of the other. The deployment analysis further suggests that the architecture remains suitable for resource-constrained robotic perception on an embedded device, with additional inference optimization required for higher real-time throughput.